\documentclass[conference]{IEEEtran}
\IEEEoverridecommandlockouts
\usepackage{cite}
\usepackage{amsmath,amssymb,amsfonts}
\usepackage{algorithmic}
\usepackage{graphicx}
\usepackage{textcomp}
\usepackage{xcolor}
\def\BibTeX{{\rm B\kern-.05em{\sc i\kern-.025em b}\kern-.08em
    T\kern-.1667em\lower.7ex\hbox{E}\kern-.125emX}}

\newcommand{\y}{\mathbf{y}} 
\newcommand{\z}{\mathbf{z}} 
\newcommand{\w}{w}
\newcommand{\uw}{\widetilde{\w}}
\newcommand{\parti}{i} 

\newcommand{\proposal}{q\left(\particle~\mid~\particle[t-1], \mathbf{y}_t\right)}
\newcommand{\lproposal}{q_\theta \left(\particle~\mid~\particle[t-1], \mathbf{y}_t\right)} 
\newcommand{\ltransition}{p_\phi\left(\particle~\mid~\particle[t-1]\right)}

\newcommand{\likelihood}{p\left(\mathbf{y}_t~\mid~\particle\right)}
\newcommand{\transition}[1][i]{p\left(\state^{#1}~\mid~\state[t-1]^{#1}\right)}

\newcommand{\state}[1][\ti]{\z_{#1}}
\newcommand{\particle}[1][\ti]{\z_{#1}^{i}}
\newcommand{\particleWeight}[1][\ti]{\w_{#1}^\parti}

\newcommand{\real}{\mathbb{R}}

\DeclareMathOperator*{\argmax}{arg\,max}

\usepackage[nolist,nohyperlinks,printonlyreused]{acronym}
\begin{acronym}
\acro{KF}{Kalman filter}
\acro{EKF}{extended Kalman filter}
\acro{UKF}{unscented Kalman filter}
\acro{DPF}{differentiable particle filter}
\acro{VSMC}{variational sequential Monte Carlo}
\acro{GMM}{Gaussian mixture model}
\acro{PSF}{point spread function}
\acro{BPF}{bootstrap particle filter}
\acro{VRNN}{variational recurrent neural network}
\acro{CRLB}{Cramér–Rao lower bound}
\acro{KL}{Kullback–Leibler}
\acro{MLP}{multilayer perceptron}
\acro{MRI}{magnetic resonance imaging}
\acro{CT}{computed tomography}
\acro{ELBO}{evidence lower bound}
\acro{SMC}{sequential Monte Carlo}
\acro{PF}{particle filter}
\acro{VAE}{variational autoencoder}
\acro{DNN}{deep neural network}
\acro{BCE}{binary cross entropy}
\acro{SGD}{stochastic gradient descent}
\acro{TPR}{true positive rate}
\acro{FPR}{false positive rate}
\acro{CNN}{convolutional neural network}
\acro{MI}{mutual information}
\acro{ViT}{visual transformer}
\acro{IoU}{intersection over union}
\acro{IIC}{invariant information clustering}
\acro{FCN}{fully convolutional network}
\acro{PhC}{phase contrast}
\acro{DIC}{differential interference contrast}
\acro{EM}{electron microscopy}
\acro{GAN}{generative adversarial network}
\acro{ROI}{regions of interest}
\acro{TIMM}{PyTorch image models}
\acro{AWGN}{additive white Gaussian noise}
\acro{PCA}{principal component analysis}
\acro{SSL}{self-supervised learning}
\acro{SAM}{segment anything model}
\acro{C.I.}{confidence interval}
\end{acronym}
\usepackage{xcolor}
\newcommand{\vsmc}{\textcolor{red}{*}}
\usepackage[ruled,vlined]{algorithm2e}

\begin{document}

\title{Deep Variational Sequential Monte Carlo for High-Dimensional Observations
{
\thanks{This work was supported by the European Research Council (ERC) under the ERC starting grant nr. 101077368 (US-ACT).}
}}

\author{\IEEEauthorblockN{Wessel L. van Nierop}
\IEEEauthorblockA{\textit{Dept. of Electrical Engineering} \\
\textit{Eindhoven University of Technology}\\
Eindhoven, The Netherlands \\
w.l.v.nierop@tue.nl}
\and
\IEEEauthorblockN{Nir Shlezinger}
\IEEEauthorblockA{\textit{Dept. of Electrical and Computer Engineering} \\
\textit{Ben-Gurion University of the Negev}\\
Beer-Sheva, Israel \\
nirshl@bgu.ac.il}
\and
\IEEEauthorblockN{Ruud J.G. van Sloun}
\IEEEauthorblockA{\textit{Dept. of Electrical Engineering} \\
\textit{Eindhoven University of Technology}\\
Eindhoven, The Netherlands \\
r.j.g.v.sloun@tue.nl}
}

\maketitle

\maketitle
\begin{abstract}
\Ac{SMC}, or particle filtering, is widely used in nonlinear state-space systems, but its performance often suffers from poorly approximated proposal and state-transition distributions. This work introduces a differentiable particle filter that leverages the unsupervised variational \ac{SMC} objective to parameterize the proposal and transition distributions with a neural network, designed to learn from high-dimensional observations. Experimental results demonstrate that our approach outperforms established baselines in tracking the challenging Lorenz attractor from high-dimensional and partial observations. Furthermore, an \acl{ELBO} based evaluation indicates that our method offers a more accurate representation of the posterior distribution.
\end{abstract}
\acresetall
\acresetall
\section{Introduction}
\label{sec:intro}
State-space estimation in dynamical systems, particularly for nonlinear systems, is a prevalent issue in statistical signal processing. Classical filters, such as the \ac{KF}~\cite{kalman_filter_1960}, compute the posterior state distribution based on prior observations, offering state point estimates and uncertainty quantification. The KF is optimal for linear Gaussian systems, with variants like the \ac{EKF}~\cite{jazwinski_stochastic_1970} and \ac{UKF}~\cite{julier_new_1997} handling nonlinear systems.

Particle filtering, or \ac{SMC}, represents the posterior state distribution of a stochastic process using particles, based on noisy or partial observations. This method accommodates nonlinear state-space models, handles arbitrary noise and initial state distributions, and enables posterior sampling without relying on strong Gaussian assumptions. However, traditional particle filters, such as the \ac{BPF}~\cite{gordon_novel_1993}, often use poorly approximated proposal distributions, making them less effective in high-dimensional systems~\cite{naesseth_elements_2022}. 
There exist particle filter variants that provide more efficient sampling in high-dimensional spaces~\cite{doucet_rao-blackwellised_2000,djuric_multiple_2007,djuric_particle_2013,bunch_approximations_2014,heng_gibbs_2020}, but they often require extensive domain knowledge.

\Acp{DPF} provide a way to propagate gradients through the normally non-differentiable resampling step~\cite{chen_overview_2023}. Consequently, this opens up the possibility of using neural networks to parameterize the probability distributions of the particle filter, in particular the state transition and proposal distribution. Corenflos et al.~\cite{corenflos_differentiable_2021} leverage entropy-regularized optimal transport to obtain a differentiable resampling scheme, showing that this form of resampling can outperform alternative differentiable resampling schemes.

Several approaches have been developed to learn the proposal distribution using \acp{DPF}~\cite{corenflos_differentiable_2021, gama_unsupervised_2023, cox_end--end_2024, chen_normalising_2024, li_learning_2023, jonschkowski_differentiable_2018, scibior_differentiable_2021}. However, these methods rely on a supervised objective, where the true state is known and the proposal distribution is optimized towards it. Such supervised learning is restricted to settings where labeled data is available, and cannot guarantee an accurate posterior distribution. In contrast, \ac{VSMC}~\cite{naesseth_variational_2018, maddison_filtering_2017, le_auto-encoding_2018} introduces an unsupervised variational objective to learn the proposal distribution of particle filters by maximizing the expected log-marginal likelihood estimate. This optimizes the particle filter using only observation data, assuming the measurement model is known and differentiable. 
This work extends the application of this \textit{unsupervised} objective to problems involving \textit{high-dimensional} observations, such as visual data. To the best of our knowledge, this approach has not previously been explored.

This paper presents the following contributions. (1) We show that the proposal and transition distributions can be learned from high-dimensional observations without ground-truth states (unsupervised) by parameterizing the distributions using neural networks and maximizing the estimated log-marginal likelihood estimate (\ac{VSMC}). (2) Experimental results using the challenging Lorenz attractor show that learning the proposal and transition distributions increases \ac{ELBO} based performance and tracking performance. (3) We show that the proposed method outperforms common alternative filtering techniques and baselines including the \ac{EKF}, \ac{BPF} and a supervised regression model.

The rest of this paper is organized as follows. Section~\ref{sec:background} provides the signal model and particle filter background, while Section~\ref{sec:methods} introduces our neural network-based methodology for proposal learning from high-dimensional observations. In Section~\ref{sec:experiments}, we present experiments conducted on high-dimensional image data from the Lorenz attractor, with the corresponding results detailed in Section~\ref{sec:results}. Finally, we conclude in Section~\ref{sec:conclusion}.

\section{Background} \label{sec:background}
\subsection{Signal model}
We consider a dynamical, possibly non-linear, continuous state-space model in discrete time $t\in \mathbb{Z}$. The state variables~$\z_t$ evolve using the evolution function~$\mathbf{f}$, given by
\begin{equation}
\label{eq:evolution}
\begin{array}{ll}
\z_t=\mathbf{f}\left(\z_{t-1}\right)+\mathbf{e}_t, & \z_t \in \real^{d_z},
\end{array}
\end{equation}
where $\mathbf{e}_t$ is the state noise. 
The observation vectors $\y_t$ are high-dimensional images and are related to the state-space through the measurement function $\mathbf{h}$, given by
\begin{equation}
\label{eq:measurement}
\begin{array}{ll}
\y_t=A_t (\mathbf{h}\left(\z_t\right)+\mathbf{v}_t), & \y_t \in \mathbb{R}^{h\times w},
\end{array}
\end{equation}
where $\mathbf{v}_t$ is the observation noise, $A_t$ is a masking operation that will randomly observe only a proportion $P$ of the image, and the observation dimensionality is $(h \times w)\gg d_z$.

\subsection{Particle filter} \label{sec:pf}
A particle filter aims to estimate the posterior density of state variables $\z_t$ given observations $\y_t$ over time~$t \in \{1,...,T\}$ using a set of $N$ weighted particles
\begin{equation}
    p(\z_t \mid \y_{1: t}) \approx \sum_{\parti=1}^{N} \w_t^{\parti} \delta \left(\z_t-\z_t^\parti\right),
\end{equation}
where $\w_t^i$ is the weight of particle $i$ and $\delta$ is the Dirac delta function. 
The particle filter used in this paper is described in Algorithm~\ref{alg:pf} and performs the resampling step when the effective sample size $\hat{N}_{eff}$ falls below a predefined threshold~$N_T$. For more information on particle filter algorithms, we refer the reader to\cite{arulampalam_tutorial_2002,naesseth_elements_2022}.

\begin{algorithm}[t]
    \caption{Particle filter with \ac{VSMC} Algorithm for a single batch of data}
    \label{alg:pf}

    $\left\{w_0^i\right\}_{i=1}^N=\{1 / N\}_{i=1}^N$

    \ForEach{$t \in \{1\ldots T\}$}{
        \ForEach{$i \in \{1\ldots N\}$}{
            Sample from proposal $\particle \sim \lproposal$\;
            Calc. weight $\uw_t^i \propto w_{t-1}^i \frac{\likelihood \ltransition}{\lproposal}$\;
        }

        \ForEach{$i \in \{1\ldots N\}$}{
            Normalize weights $\particleWeight = \frac{\uw_t^{i}}{\sum_{j=1}^N \widetilde{w}_t^{j}}$\;
        }
        Compute $\hat{N}_{eff}=\left(\sum_{i=1}^{N}\left(w_t^i\right)^2\right)^{-1}$\;
        \If{$\hat{N}_{eff} < N_T$}{
            \ForEach{$i \in \{1\ldots N\}$}{
                $\particle \sim \textsc{Resample}^{\color{blue}{\dagger}} (\{\particle, \particleWeight\})$;\hspace{0.1cm}
                $w_t^i = \frac{1}{N}$\;
            }
        }

        \vsmc Compute $\hat{p}(\y_t \mid \y_{1: t-1})=\frac{1}{N} \sum_{i=1}^N \uw_t^i$\;
        \vspace{1mm}
        
        \vsmc Aggregate $\log \hat{p}(\y_{1:t}) = \log \hat{p}(\y_{1:t-1})+\log \hat{p}(\y_t \mid \y_{1: t-1})$\;
    }
    \vsmc Optimize $\argmax_\theta \log \hat{p}(\y_{1:T})$;

    \vspace{0.2cm}
    \vsmc is part of \ac{VSMC}\\
    $^{\color{blue}{\dagger}}$differentiable resampling during training and systematic resampling during inference.
\end{algorithm}

The \ac{BPF} uses $\proposal = \transition$ as proposal distribution, and therefore does not use observations to propose new particles. This also means that the weights of the particle filter are determined only by the likelihood~$\likelihood$ because the proposal and transition probabilities cancel out. However, the observations are invaluable input for the proposal distribution, particularly when the state-evolution model is not accurate or hard to estimate.

\begin{figure}[t]
    \centering
    \includegraphics[width=\linewidth]{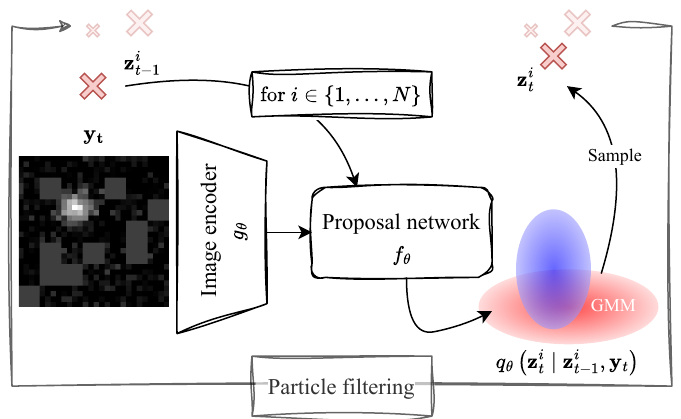}
    \caption{Illustration demonstrating the proposed method of using a parameterized proposal distribution in particle filtering.}
    \label{fig:diagram}
\end{figure}

\section{Methods} \label{sec:methods}
\subsection{Proposal learning} \label{sec:proposal}
To overcome the shortcomings of the \ac{BPF}, we learn the proposal distribution using the unsupervised \ac{VSMC} objective. The objective function for \ac{VSMC}~\cite{naesseth_variational_2018, le_auto-encoding_2018, maddison_filtering_2017} is the expected \ac{SMC} log-marginal likelihood estimate
\begin{equation}
\label{eq:objective}
    \mathbb{E}\left[\log \hat{p}\left(\y_{1: T}\right)\right] = \sum_{t=1}^T \log \left(\frac{1}{N}\sum_{i=1}^N \uw_t^i\right),
\end{equation}
which is proven to be a suitable variational objective and a lower bound to the true \ac{ELBO} \cite{naesseth_variational_2018}.

We use \acp{GMM} for the proposal distribution
\begin{equation}
    \lproposal=\sum_{k=1}^K \pi_k^i \mathcal{N}\left(\boldsymbol{\mu}_k^i, \boldsymbol{\sigma}_k^i I \right),
\end{equation}
where $K$ represents the number of mixture components. The \acp{GMM} are parameterized by a neural network~$\theta$, with parameters~$\theta$ learned from data using the differentiable resampling scheme of~\cite{corenflos_differentiable_2021}. At inference time, the faster systematic resampling scheme~\cite{carpenter_improved_1999} can be used. We select \acp{GMM} for their ability to model multi-modality and their well-defined reparameterization trick~\cite{graves_stochastic_2016, figurnov_implicit_2019}. The parameterized proposal distribution is optimized via stochastic gradient ascent on \eqref{eq:objective} using the AdamW optimizer~\cite{loshchilov_decoupled_2019}, as described in the optimization procedure of Algorithm~\ref{alg:pf}. To avoid numerical errors from small log-likelihood values, the sequence length $T$ is gradually increased during training~\cite{cox_end--end_2024}.

\subsection{Networks} \label{sec:network}
The proposal distribution $q_\theta$ is formed by two neural networks. First, a convolutional image encoder $g_\theta: \real^{h\times w} \rightarrow \real^{d_c}$ will compress the observation image into an encoded representation. Second, a \ac{MLP} $f_\theta: \real^{d_z + d_c} \rightarrow \real^{K  (2 d_z +  1)}$ takes the image encoding~$g_\theta (\y_t)$ and outputs, for every particle, the parameters of a \ac{GMM} as given by
\begin{equation}
    \left\{ (\pi_1^i, \boldsymbol{\mu}_1^i, \boldsymbol{\sigma}_1^i), \ldots, (\pi_K^i, \boldsymbol{\mu}_K^i, \boldsymbol{\sigma}_K^i)\right\} = f_\theta \left(\particle[t-1], g_\theta (\y_t)\right).
\end{equation}
Fig.~\ref{fig:diagram} illustrates the application of the neural proposal distribution.
The convolution image encoder consists of three convolutional layers with a ReLU activation function. The number of convolutional features increases by a factor of 2 at each layer, while the image dimensions shrink by a factor of 2 using max pooling. Finally, a single fully connected layer maps the filters to an encoding of $d_c=256$ parameters.
The rest of the proposal network consists of a six-layer \ac{MLP} of 256 dimensions with ReLU activation. The final layer maps to a vector of size $K  (2 d_z +  1)$, which corresponds to the mean, log-variance, and mixture weights of the \ac{GMM}.

The transition distribution is jointly optimized and parameterized by an equivalent network architecture as the aforementioned \ac{MLP} but does not take observations as input such that $f_\phi: \real^{d_z} \rightarrow \real^{K  (2 d_z +  1)}$.

\section{Experiments} \label{sec:experiments}

\subsection{Lorenz attractor} \label{sec:lorenz}
\begin{figure}
    \centering
    \includegraphics[width=\linewidth]{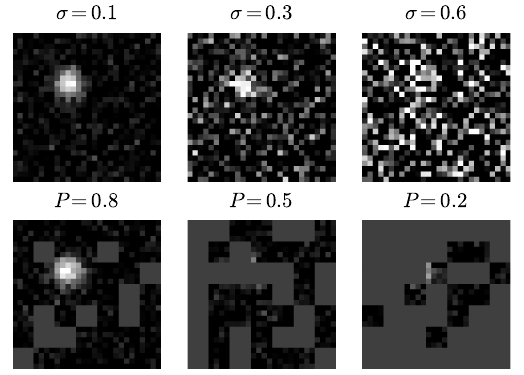}
    \caption{Example images of the Lorenz attractor (in the same position) using various noise levels with standard deviation~$\sigma$ (top) and randomly dropped observations where $P$ represents the proportion of observations (bottom).}
    \label{fig:lorenz}
\end{figure}

To demonstrate the applicability of the \ac{DPF} model for high-dimensional observations, we use images generated from the Lorenz attractor~\cite{lorenz_deterministic_1963}, which exhibits chaotic nonlinear movement in 3-D space. 
We follow~\cite{buchnik_latent-kalmannet_2023, garcia_satorras_combining_2019}
in projecting the state to an image using a Gaussian \ac{PSF} as measurement function $\mathbf{h}$.
The observations $\y_t$ are images of 28$\times$28 pixels.
For noise distributions $\mathbf{e}_t$ and $\mathbf{v}_t$ we choose \ac{AWGN} with a standard deviation of 0.5 for $\mathbf{e}_t$ and a standard deviation in the range of [0.1, 0.6] for $\mathbf{v}_t$. Here, $A_t$ in \eqref{eq:measurement} is the identity function.
Additionally, we conduct an experiment where the images are partially observed by randomly dropping 4$\times$4 pixel blocks with varying probabilities, as defined by $A_t$. In this experiment, \ac{AWGN} is fixed to a standard deviation of 0.1.
Some examples can be seen in Fig.~\ref{fig:lorenz}.

The measurement model $\likelihood$ maps the particles from the state space to the observation space using the \ac{PSF}, and then calculates the pixel-wise error assuming a Gaussian distribution, with variance corresponding to the AWGN present in the images.
In our experiments, the transition distribution $\transition$ is considered unknown, leading to a mismatch with the simulated state transitions and adding to the complexity of the problem.

\subsection{Baselines} \label{sec:baselines}
The proposed unsupervised model was compared to several baselines, including the \ac{EKF}~\cite{jazwinski_stochastic_1970}, \ac{BPF}~\cite{gordon_novel_1993}, and a supervised neural network.
The supervised network was trained to map noisy or partially observed images to an estimated coordinate using Euclidean distance as the loss function. This non-variational encoder estimates the posterior mean coordinate. It mirrors the network for the proposal distribution in Section~\ref{sec:network}, but the final layer outputs a coordinate estimate instead of distribution parameters.

The state-space dynamics are considered to be unknown in this work. To address this, the \ac{BPF} and \ac{EKF} estimate velocity as part of the state and assume a constant velocity transition model. In contrast, the DPF simultaneously learns the transition and proposal distribution, allowing it to adapt to complex dynamics. Unlike the baselines, which initialize particles around the initial position using a Gaussian distribution with unit variance, the DPF initializes particles using an estimated prior distribution. This makes the DPF significantly more flexible, as it does not rely on a predefined initial position. All particle filters use $N=28$ particles except for the \ac{BPF} with 280 particles. The resample threshold is always~$N_T = \frac{N}{2}$. Validation is performed on 32 unseen sequences of 128 timesteps while the models are trained on 1024 sequences of 8 timesteps divided into batches of 32 sequences. We trained the \ac{DPF} and the supervised encoder once for multiple noise levels and once for the various levels of partial observations.
The DPF employs a proposal and transition distribution with $K=2$ mixture components.

\begin{figure}[t]
    \centering
    \includegraphics[width=\linewidth]{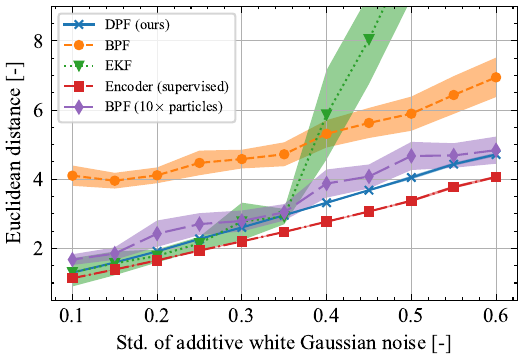}
    \caption{Tracking error of the \ac{DPF} compared to the baseline methods for varying noise levels. The tracking error is computed using the Euclidean distance of the predicted location to the ground truth location. The error bars depict the 95\% confidence interval calculated using 20 different seeds.}
    \label{fig:tracking_error_noise}
\end{figure}

\subsection{Metrics} \label{sec:metrics}
We evaluate the accuracy of the state estimate in terms of Euclidean distance as well as the \ac{ELBO} for the various filtering techniques. The \ac{ELBO} can be decomposed into the likelihood of the observations and the \ac{KL} divergence to the prior as given by
\begin{equation}
\begin{split}
        \ln p(\y_t) & \geq \mathbb{E}_{\z_t \sim q(\cdot \mid \y_t)}[\ln p(\y_t \mid \z_t)] \\ 
        & -\mathbb{E}_{\z_t \sim q(\cdot \mid \y_t)}\left[\ln q(\z_t \mid \y_t)-\ln p(\z_t)\right].
\end{split}
\end{equation}
The prior $p(\z)$ is estimated by aggregating many timesteps of the Lorenz attractor and using Gaussian kernel density estimation. The particles from the \ac{SMC} posterior are fitted with a Gaussian to enable sampling and evaluating arbitrary points. The likelihood and \ac{KL} are estimated using Monte Carlo sampling.

In addition to evaluating the variational performance, the tracking performance is assessed using the posterior mean. The posterior mean is determined by calculating the weighted mean of the particles. Following this, the Euclidean distance is measured relative to the ground truth coordinates of the attractor.

\begin{figure}
    \centering
    \includegraphics[width=\linewidth]{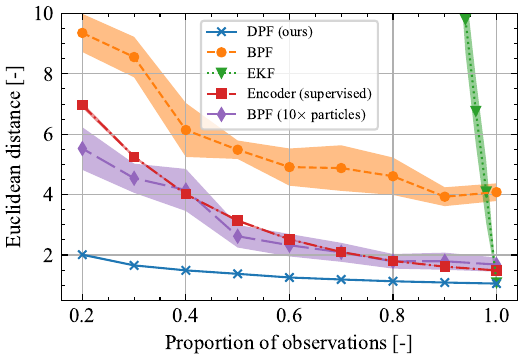}
    \caption{Tracking error of the \ac{DPF} compared to the baseline methods for various levels of partial observations. The tracking error is computed using the Euclidean distance of the predicted location to the ground truth location.  The error bars depict the 95\% confidence interval calculated using 10 different seeds.}
    \label{fig:tracking_error_partial}
\end{figure}

\begin{figure}
    \centering
    \includegraphics[width=\linewidth]{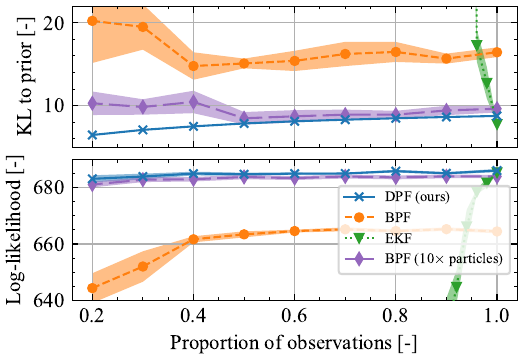}
    \caption{Decomposition of the \ac{ELBO} for the \ac{DPF} and the baselines for various levels of partial observations. The error bars depict the 95\% confidence interval calculated using 10 different seeds.}
    \label{fig:elbo}
\end{figure}

\section{Results} \label{sec:results}
Fig.~\ref{fig:tracking_error_noise} shows the tracking error as a function of the \ac{AWGN} standard deviation. As anticipated, the supervised method achieves the best performance due to the fully observed image. Among unsupervised models, the \ac{DPF} performs the best. Up to a standard deviation of 0.35, the \ac{EKF} performs similarly to the \ac{DPF}, but its error rises sharply beyond this point. Meanwhile, the \ac{BPF} demonstrates stable but suboptimal performance, even when using 10 times the number of particles.

Tracking error under varying levels of dropped observations is illustrated in Fig.~\ref{fig:tracking_error_partial}. The \ac{DPF} significantly outperforms all other methods, with the supervised encoder and \ac{BPF} (10$\times$ particles) being its closest competitors. With limited observations, the supervised encoder deviates from \ac{DPF} due to the lack of a state-transition model. The \ac{EKF} struggles with partial observations.

The ELBO decomposition in Fig.~\ref{fig:elbo} reveals that the \ac{DPF} reduces the \ac{KL} divergence to the prior as observation uncertainty increases, while maintaining a stable log-likelihood. In contrast, the \ac{BPF} with the same number of particles remains relatively stable, but drops sharply when observations fall below 40\%, and the \ac{EKF} deviates significantly. The figure demonstrates that the \ac{DPF} provides the most accurate modeling of the posterior distribution.

\section{Conclusion} \label{sec:conclusion}
We proposed an unsupervised neural augmentation of particle filters for dynamical state estimation from high-dimensional observations. Our approach leverages parameterized proposal and transition distributions and the unsupervised \ac{VSMC} objective to effectively mitigate the challenge of poorly approximated proposals. It learns from high-dimensional observations without the need for ground-truth state information, showing significant improvements over established baselines on the challenging Lorenz attractor, especially in cases with limited observations. Additionally, the ELBO-based evaluation confirms that the posterior distribution is modeled more accurately.

\bibliographystyle{IEEEtran}
\bibliography{IEEEabrv,refs2}

\end{document}